\documentclass{article}
\usepackage[utf8]{inputenc}
\usepackage{shi_dong_template}
\usepackage{algorithmic}
\usepackage[numbers]{natbib}
\usepackage{comment}
\newcommand{\cmmnt}[1]{\ignorespaces}

\title{Provably Efficient Reinforcement Learning\\ with Aggregated States}
\date{}
\author[1]{Shi Dong}
\author[1]{Benjamin Van Roy}
\author[2]{Zhengyuan Zhou}
\affil[1]{\small Stanford University}
\affil[2]{\small New York University}

\def\C#1{\mathcal{#1}}

\def\BB#1{\mathbb{#1}}

\def\a{\alpha}
\def\b{\beta}
\def\d{\delta}

\def\e{\epsilon}

\def\nn{\nonumber}

\def\E{\mathbb{E}}
\def\h#1{\hat{#1}}

\def\t#1{\tilde{#1}}

\def\state{\mathcal{S}}
\def\action{\mathcal{A}}
\def\tran{\mathrm{P}}

\def\ind{\mathbf{1}}
\def\regret{\mathrm{Regret}}
\def\tQ{\tilde{Q}}
\def\hQ{\hat{Q}}

\def\hV{\hat{V}}
\def\tO{\tilde{\mathcal{O}}}
\def\dn{\check{n}}

\def\htau{\hat{\tau}}
\def\hn{\hat{n}}

\def\algname{\texttt{AQ-UCB}}
\def\nn{\nonumber}

\renewenvironment{equation*}
{\begin{eqnarray}}{\end{eqnarray}\ignorespacesafterend}

\begin{document}
\maketitle

\begin{abstract} 
We establish that an optimistic variant of Q-learning applied to a fixed-horizon episodic Markov decision process with an aggregated state representation incurs regret $\tO(\sqrt{H^5 M K} + \e HK)$, where $H$ is the horizon, $M$ is the number of aggregate states, $K$ is the number of episodes, and $\e$ is the largest difference between any pair of optimal state-action values associated with a common aggregate state.  Notably, this regret bound does not depend on the number of states or actions and indicates that asymptotic per-period regret is no greater than $\epsilon$, independent of horizon.  To our knowledge, this is the first such result that applies to reinforcement learning with nontrivial value function approximation without any restrictions on transition probabilities.
\end{abstract}

\section{Introduction}

Value function learning with aggregated states serves as a foundational subject of reinforcement learning (RL).  With such a representation, the set of all state-action pairs is partitioned, with each cell representing an {\it aggregate state}.  An agent learns an approximation to the state-action value function for which value is constant across each cell.  In this paper, we consider an optimistic version of Q-learning that applies with an aggregated state representation.  We show that the algorithm enjoys an $\tO(\sqrt{H^5 M K} + \e HK)$ worst-case regret bound, where $H$ is the horizon, $M$ is the number of aggregate states, $K$ is the number of episodes, and $\e$ is the maximum difference between optimal state-action values associated with a common aggregate state. Notably, our result does not depend on any structural assumption on the environment.  Both the algorithm and the analysis build on the recent work of \citet{jin2018q}, which restricted attention to tabular representations.

A large and growing body of works establish guarantees for RL algorithms with tabular representations (see, e.g., \cite{strehl2006pac,jaksch2010near,azar2011speedy,osband2013more,jin2018q,osband2019deep,russo2019worst,agarwal2019optimality}).  Data and thus learning time required with tabular representations typically grow with the number of state-action pairs.  For problems of practical scale, this renders tabular representations impractical.  State aggregation offers an approach to reducing this statistical complexity.

A number of prior results imply bounds on learning time with aggregated states, as does the regret bound we will establish.  However, these prior results impose requirements on the structure of the environment, while we do not.
For example, \citet{wen2017efficient} develop an algorithm for deterministic environments and establish efficient learning with aggregated states.
Another line of recent work \cite{jin2019provably,yang2019reinforcement,zanette2019frequentist} establishes efficiency guarantees that apply with general linearly parameterized value functions, of which state aggregation representations constitute a special case.  However, these results rely on transition kernels being specially structured, obeying a low-rank property.
\citet{jiang2017contextual} establish sample efficiency of an algorithm that can work with generalizing representations like state aggregators.  The result requires small ``Bellman rank,'' which holds when $\epsilon = 0$ but may not when $\epsilon > 0$.

Another important consideration when learning with aggregated state representations is the ultimate performance of the agent.  The performance typically will not converge on optimal if $\epsilon > 0$, and different algorithms may lead to different eventual performance.  Our regret bound establishes that eventual performance loss is no greater than $O(\epsilon)$ per period.
Interestingly, results of \citet{lagoudakis2003least,li2009unifying,yang2019theoretical,zanette2019frequentist}
pertaining to alternative reinforcement learning algorithms that can be applied with aggregated states point to
eventual loss of $O(\epsilon H)$ per period.
The algorithm of \citet{wen2017efficient}, which only applies to deterministic environments, 
leads to $O(\epsilon)$ per period performance loss, but only if the algorithm converges on a fixed policy, which is not guaranteed.
This difference between algorithms that achieve $O(\epsilon)$ versus $O(\epsilon H)$ per period loss has been studied in the approximate dynamic programming context by 
 \citet{van2006performance}, who identified properties of temporal-difference methods that yield the former.  Performance bounds for other approximate dynamic programming methods exhibit the $O(\epsilon H)$ dependence, as appears, for example, in \cite{Gordon95,Tsitsiklis96,munos2008finite}.

\section{Problem Formulation}

We consider an agent sequentially interacting with an environment with state space $\state$ and action space $\action$. We assume that both spaces are finite, with cardinality $S$ and $A$, respectively.  Each episode of interaction consists of $H$ stages, and produces a sequence
\[
    s_1, a_1,\dots, s_H, a_H,
\]
where for $h=1,\dots,H$, $s_h\in\state$ is the system state in which the agent resides at the beginning of stage $h$, and $a_h\in\action$ is the action taken by the agent after $s_h$ is observed. For simplicity, we assume that at the beginning of each episode, the system is reset to a deterministic state $s_1$. The dynamics of the system is governed by the transition kernels $\tran_h$, which specify the distribution of the next state, given the current system state and the action that the agent takes, i.e.
\[
    \tran_h^{s,a}(s') = \mathbb{P} \left(s_{h+1} = s'\ \big|\ s_h=s, a_h=a\right).
\]
A deterministic reward\footnote{We assume that the rewards are deterministic only to streamline the analysis. All our results apply without change to environments with stochastic but bounded rewards.} $R_h(s,a)$ is associated with each state-action pair $(s,a)$ at stage $h$. We assume that rewards take values in $[0, 1]$. At the final stage $H$, the episode terminates after the agent takes $a_H$ in state $s_H$, and realizes the reward $R_H(s_H, a_H)$. The goal of the agent is to maximize the total reward accrued in an episode, namely the {\it episodic return}, defined as $\sum_{h=1}^H R_h(s_h, a_h)$.\par 

At each stage, a learning algorithm prescribes a specific distribution over $\action$, from which the agent draws the next action.  Such a sequential prescription is called a {\it policy}.  The policy is said to be  deterministic if each distribution is concentrated on one single action. In this work, we only consider deterministic policies, which can be concretized as mappings from $\state$ to $\action$. We say that the agent follows policy $\pi$, if for all $h\in[H]$\footnote{For positive integer $N$, we use the notation $[N]$ as a shorthand for $\{1,2,\dots,N\}$.},
\[
    a_h = \pi_h(s_h).
\]
The {\it value function} of policy $\pi$ is defined as the expected return realized by the agent when she follows policy $\pi$, namely
\[
    V_h^\pi(s) = \E\left[ 
        \sum_{i=h}^H R_i(s_i, \pi_i(s_i)) \Big|
        s_h = s
    \right],
\]
where the expectation is taken over all possible transitions. 
For each policy $\pi$, we can also define the {\it state-action value function} or {\it Q-function} of state-action pair $(s,a)$ as the expected return when the agent takes action $a$ at state $s$, and then follows policy $\pi$, i.e.
\[
    Q_h^\pi(s,a) = R_h(s, a) + \E\left[ 
        	\sum_{i=h+1}^H R_i(s_i, \pi_i(s_i)) \Big|
        	s_h = s, a_h = a
    	\right].\nn
\]
There exists an optimal policy $\pi^*$, which is deterministic, and maximizes $V_h^{\pi}(s)$ for every $s$. We denote by $V^*$ the value function corresponding to the optimal policy, and define 
\[
    Q_h^*(s,a) = R_h(s, a) + \tran_h V^*_{h+1}(s,a),
\]
where we use the notation 
\[
    \tran_h V(s,a) = \E_{s'\sim \tran_h^{s,a}}[V(s')].
\]
Notice that $Q^*_h(s,a)$ is the maximum realizable expected return when the agent starts from $s$ and takes action $a$ at stage $h$.  Therefore we also call it the ``ground-truth'' value of the state-action pair $(s,a)$. 
From the optimality of $V^*$, we have that
\[
    V_h^*(s) = \max_{a\in\action} Q_h^*(s,a),\quad \forall h \in [H],\ s\in\state.
\]
It is worth noting that under our bounded rewards assumption, we have that
\[
    0 \leq V_h^\pi \leq V^*_h \leq H,
\]
for all $h=1,\dots, H$ and policy $\pi$.\par 

The goal of an RL algorithm is to identify a good policy through consecutive interactions with the environment, with no prior knowledge of environment dynamics $\tran$ and reward $R$.  We will evaluate algorithm performance in terms of {\it cumulative regret}.  Formally, taking $\pi_1,\dots, \pi_K$ to be the sequence of policies generated by the algorithm over $K$ episodes, the cumulative regret is defined by
\[
    \regret(K) = \sum_{k=1}^K V_1^*(s_1) - V_1^{\pi_k}(s_1).
\]

In a value function learning algorithm, the goal is to learn an approximation of the value function from which an effective policy can be derived.  However, if we aim to separately estimate the value of each state-action pair, for example, using a tabular representation, the data and computational requirements will scale at least linearly in $SA$, which is infeasible for problems of practical scale.
Aggregated state representations reduce complexity and accelerate learning by aggregating state-action pairs.  This involves partitioning the set of state-action pairs into $M$ cells.  Each cell can be thought of as an aggregate state, and our value function representation only has to maintain one value estimate per aggregate state.  
Let $\Phi$ be the set of aggregate states, and  $\phi_h:\state\times\action\mapsto\Phi$ be the mapping from state-action pairs to aggregate states at stage $h$. Without loss of generality, we let $\Phi=[M]$.  Formally, we can define an aggregated state representation as follows.
\begin{definition}
    \label{def:max-span}
    We say that $\{\phi_h\}_{h=1}^H$ is an $\e$-error aggregated state representation (or $\e$-error aggregation as a shorthand) of an MDP, if for all $s, s'\in\state,\ a, a'\in\action$ and $h\in[H]$ such that $\phi_h(s,a) = \phi_h(s',a')$,
    \[
        \left|Q^*_h(s, a) - Q^*_h(s', a')\right| \leq \e.
    \]
\end{definition}

An algorithm with aggregated state representation $\{\phi_h\}_{h=1}^H$ posits that the ground-truth Q-function lies in or close to the hypothesis function class $\C{F}_1\times\dots\times\C{F}_H$, where
\[
	\C{F}_h =\Big\{f:\state\times\action\mapsto\BB{R}\ \Big|\ f(s, a) = f(s', a') \text{ if } \phi_h(s, a) = \phi_h(s', a') \Big\},\ h\in[H]. \nn
\]

If $\{\phi_h\}_{h=1}^H$ is a $0$-error aggregation of an MDP, then we say that the aggregation is {\it sufficient}, in the sense that the value of an aggregate state exactly represents the values of all state-action pairs mapped to it. This corresponds to the case where the actual value function lies in the hypothesis function class. As expected, only under this case can we guarantee that an algorithm finds the optimal policy as $K\to\infty$.  When $\e>0$ in Definition \ref{def:max-span}, there exists an MDP such that no RL algorithm with aggregated state representation $\{\phi_h\}_{h=1}^H$ can find the optimal policy \cite{van2006performance}.  In this case, the best we can hope for is that the algorithm finds a policy whose suboptimality can be upper bounded by a function of $\e$.

\section{Algorithm}
\label{sec:alg}
In this section we present the algorithm Aggregated Q-learning with Upper Confidence Bounds (\algname), which is an optimistic version of Q-learning with an aggregated state representation. The algorithm maintains a sequence of Q-function estimates $\{\h{Q}_h\}_{h\in[H]}$.  Since the state-action pairs mapped to the same aggregate state are not distinguished in the algorithm, each estimate $\h{Q}_h$ is a mapping from $\Phi$ to real values.  For each state-action pair $(s,a)\in\state\times\action$, $\hQ_h(\phi_h(s, a))$ is the estimated value of $(s,a)$ at stage $h$.  A detailed description of \algname\ is given in Algorithm 1.

\begin{algorithm}[tb]
    \label{alg:base}
    \caption{\algname}
    \begin{algorithmic}[1]
	    \STATE {\bf Input:} $\state, \action, H, \{\phi_h\}_{h=1}^H, s_1, K$
	    \STATE {\bf Input:} positive constants $\{\b_n\}_{n=1,2,\dots}$
	    \STATE Define constants $\a_t \leftarrow (H+1)/(H+t),\ t=1, 2, \dots$
	    \STATE Initialize $N_h(m) = 0, \hQ_h(m) = H$ for all $h\in[H]$ and $m\in[M]$
	    \STATE Randomly draw the first trajectory $s_1^0, a_1^0, \dots, s_H^0, a_H^0$, where $s_1^0=s_1$
	    \FOR {{\rm episode} $k=1,\dots, K$}
	        \FOR {{\rm stage} $h=1,\dots, H$}
                \STATE $m \leftarrow \phi_h(s_h^{k-1}, a_h^{k-1})$
                \STATE $N_h(m) \leftarrow N_h(m) + 1$
                \STATE $\h{V}_{h+1} \leftarrow \max_{a\in\action} \hQ_{h+1}(s_{h+1}^{k-1}, a)$ 
                \STATE $\t{Q}_h(m) \leftarrow (1 - \a_{N_h(m)}) \cdot \h{Q}_h(m) + \a_{N_h(m)} \cdot \left[ r_h^{k-1} + \h{V}_{h+1} + \b_{N_h(m)}\cdot \frac{1}{\sqrt{N_h(m)}}\right] $\label{alg:update}
                \STATE $\h{Q}_h(m) \leftarrow \min\left\{\t{Q}_h(m),\ H \right\}$ \label{alg:capping}
	        \ENDFOR
	        \STATE $s_1^k\leftarrow s_1$
	        \FOR {{\rm stage} $h=1,\dots, H$}
	            \STATE Take action $a_h^k \leftarrow \argmax_{a\in\action} \h{Q}_h\big(\phi_h(s_h^k, a)\big)$
	            \STATE receive reward $r_h^k$ and next state $s_{h+1}^k$
	        \ENDFOR
	    \ENDFOR
        \STATE {\bf Output:} the greedy policy with respect to $\{\hQ_h\}_{h\in[H]}$
    \end{algorithmic}
\end{algorithm}\par

Among the inputs, $\{\beta_n\}$ is a sequence of positive real numbers controlling the amount of optimism injected into the value function estimates during each update.  Line 11 \cmmnt{COUNTER ALERT} in Algorithm 1 corresponds to the standard Q-learning update, with an extra term of optimistic boost:
\[
    \hQ_h(s,a) 
    &\leftarrow& (1 - \a)\cdot \hQ_h(s,a) + \nn\\
        &&\a\cdot\left[ R_h(s,a) + \max_{a'\in\action}\hQ_{h+1}(s', a') + B \right],\nn
\]
where $s'$ is the observed next state after taking $a$ at state $s$, $\a$ is the learning rate, and $B$ is the optimistic boost term aiming to direct the agent to the under-explored part of the state space.  In Line 12\cmmnt{COUNTER ALERT}, we truncate the value estimates by $H$, to prevent the estimates from being overly optimistic (recall that the ground-truth value function is upper bounded by $H$). In each episode, after updating the value estimates, the algorithm samples a new trajectory using the greedy policy with respect to the current value estimates (Lines 14-18\cmmnt{COUNTER ALERT}).  Note that, since action $a_{h+1}^{k-1}$ is greedy with respect to $\hQ_{h+1}$, it is the action that attains the maximum in Line 10\cmmnt{COUNTER ALERT}.  Therefore, without the optimistic boosting and the value estimate capping, \algname\ amounts to SARSA with actions chosen greedily.

It is also worth noting that our choice of stepsizes 
\[
	\a_t = \frac{H+1}{H+t}
\]
is the same as in \cite{jin2018q}.  This specific sequence of stepsizes puts more weight on recent updates, and enjoys many desirable properties (Lemma 4.1 in \cite{jin2018q}), thus preventing approximation errors from propagating in an exponential manner.  Other choices of stepsizes, for example $\h{\a}_t = 1/t$, which amounts to taking the average of all previous updates, may result in a learning time exponential in the horizon.

The algorithm scans through the newly obtained trajectory and updates the value function estimates in each episode.  It only has to maintain the values of $\{\h{Q}_h\}_{h\in[H]}$ and $\{N_h\}_{h\in[H]}$ throughout its execution.  Therefore, assuming that the computation of $\phi_h(s,a)$ takes $\C{O}(1)$ time, the time complexity of \algname\ is $\C{O}(HMK + HAK)$, and the space complexity is $\C{O}(HM)$.

\section{Main Results}
\label{sec:result}
The worst-case efficiency guarantee of \algname~is given in the following theorem.

\begin{theorem}
    \label{thm:main}
    Suppose $\{\phi_h\}_{h\in[H]}$ is an $\e$-error aggregation of the underlying MDP.  We have that, for any $\d>0$, if we run $K$ episodes of algorithm \algname~with
    \[
        \b_i = 2H^{\frac{3}{2}}\sqrt{\log\frac{HK}{\d}} + \e\cdot \sqrt{i},\quad i=1,2\dots,
    \]
    then with probability at least $1-\d$, 
    \[
        \regret(K) 
        &\leq& 24\sqrt{H^5MK\log\frac{3HK}{\d}}  \nn\\
        	&&+\ 12\sqrt{2H^3K\log\frac{3}{\d}}  \nn\\
        	&&+\ 3H^2M + 6\e\cdot HK.
    \]
\end{theorem}
Note that Theorem \ref{thm:main} does not rely on any assumption with respect to the environment itself. The only requirement is that the aggregated state representation $\{\phi_h\}_{h\in[H]}$ approximately captures the structure of the ground-truth value function.  In practice, such a representation is often designed by incorporating intuitions and knowledge toward the environment.

When the aggregation is sufficient, i.e. $\e=0$, Theorem \ref{thm:main} shows that the cumulative regret of \algname~is $\tO(\sqrt{H^5MK})$, which translates into $\tO(\sqrt{H^4MT})$ if we let $T=HK$ be the total number of learning periods of the agent.  When applied to the tabular representation where $M=SA$, the bound matches the Hoeffding-type regret bound in \cite{jin2018q}. The result on linear MDPs (Theorem 3.1 in \cite{jin2019provably}) also implies a regret bound $\tO(\sqrt{H^3M^3T})$ for aggregated state representations.  However, the linear MDP assumption (Assumption A in \cite{jin2019provably}) is equivalent to the assumption that the aggregate states are efficient for {\it all} policies of the MDP, i.e.
\[
	\phi_h(s, a) = \phi_h(s', a')\Rightarrow Q^\pi(s,a) = Q^\pi(s', a'), \forall \pi,
\]
whereas our Thoerem \ref{thm:main} only need the aggregate states to be efficient for the optimal policy, which is much less restrictive and easier to verify in practice.

In the case with model misspecification where $\e>0$, the regret bound in Theorem \ref{thm:main} has an extra term $\C{O}(\e HK)$, which shows that the per period performance loss of the policy that \algname~ultimately outputs is $\C{O}(\e)$. This result matches the per period loss lower bound $\Omega(\e)$ established in \cite{van2006performance}.  In fact, many commonly applied algorithms lead to a worst-case per period loss $\Omega(\e H)$, for example, as is shown in \cite{vanroy19value}, if we use a replay buffer to store the trajectories from the past episodes, and uniformly sample trajectories from the replay buffer to update the Q-function estimates.  A number of existing results, such as those in \citet{lagoudakis2003least,li2009unifying,yang2019theoretical,zanette2019frequentist}, can also only establish $\C{O}(\e H)$ loss per period.

Finally, under our episodic setting with an aggregated state representation, we can also show the following regret lower bound, which is a direct implication of Theorem 3 in \cite{jin2018q}.

\begin{theorem}
    There exists a problem instance and a $0$-error aggregation with $M$ aggregate states, such that the expected cumulative regret of any algorithm is $\Omega(\sqrt{H^3MK})$.
\end{theorem}
\proof  From Theorem 3 in \cite{jin2018q}, there exists an episodic MDP instance with $S$ states, $A$ actions and horizon $H$, such that the expected regret of any learning algorithm is at least $\Omega(\sqrt{H^2SAT})$. Consider the aggregated state representation that assigns each state-action pair to an individual aggregate state at each stage, with $M=SA$ aggregate states per stage. Apparently such an aggregation is $0$-error, and any learning algorithm has to incur $\Omega(\sqrt{H^3MK})$ regret in $K$ episodes. \qed

\section{Proof Outline of Theorem \ref{thm:main}}
In this section we briefly outline the ideas to establish the efficiency of \algname.  For simplicity, we focus on the case where $\e=0$, i.e. the aggregation is sufficient.  A complete proof of Theorem \ref{thm:main} can be found in the supplementary material.

To distinguish the value estimates at the end of each episode, we will use the following set of notations:
\begin{itemize}
	\item $\hQ_h^k(m)$: the value function estimate $\hQ_h$ of aggregate state $m$, at the end of episode $k$, with $\hQ_h^0(m) = H$.
	\item $\tQ_h^k(m)$: the uncapped value function estimate $\tQ_h$ of aggregate state $m$, at the end of episode $k$.  We have that
	\[
		\hQ_h^k(m) = \min\left\{\tQ_h^k(m), H \right\}.
	\]
	\item $N_h^k(m)$: the number of visits to aggregate state $m$ at stage $h$ in the first $k$ trajectories (indexed from 0 to $k-1$).
	\item $\tau_h^j(m)$: the episode index of the $j$-th visit to aggregate state $m$, at stage $h$. There should be
    \[
        \phi_h(s_h^{\tau_h^j(m)}, a_h^{\tau_h^j(m)}) = m,\quad \forall j=1,2\dots
    \]
	\item $\pi_k$: the greedy policy with respect to $\{\hQ_h^k(m)\}$, i.e. the policy that the agent follows to produce the trajectory $s_1^k, a_1^k, s_2^k, \dots, s_H^k, a_H^k$.  The final policy output by the algorithm is $\pi_K$.
	\item $\hV_h^k(s)$: the state value function estimate at stage $h$, induced by $\hQ_h^k(m)$ through
	\[
		\hV_h^k(s) = \max_{a\in\action} \hQ_h^k(\phi_h(s, a)).\nn
	\]
\end{itemize}
To make notations concise, we will simplify $\hQ_h^k(\phi_h(s, a))$, $\tQ_h^k(\phi_h(s, a))$ and $N_h^k(\phi_h(s, a))$ as $\hQ_h^k(s, a)$, $\tQ_h^k(s, a)$ and $N_h^k(s, a)$, respectively.  Since here we consider $\e=0$, all $\b_i$ have the same value and therefore we can drop the subscripts.  Also we will use $Q^*_h(m)$ to denote the ground-truth value of all state-action pairs mapped to aggregate state $m$ at stage $h$.   Our proof consists of three parts.  The first part shows that the value function estimates in \algname~are optimistic with high probability.  The second part demonstrates that they are not ``overly optimistic,'' in the sense that there exists a polynomial upper bound for the sum of on-policy errors across all episodes.  The final part concludes the proof.

\subsection{Optimism}
Adopting the notations
\[
	\a_t^0 = \prod_{j=1}^t (1 - \a_j)
\]
and 
\[
	\a_t^i = \a_i\prod_{j=i + 1}^t (1 - \a_j),
\]
the uncapped value function estimates, computed in Line 11 \cmmnt{COUNTER ALERT} of Algorithm 1, can be written as
\begin{align*}
\tQ_h^k(m)
=\a_{N_h^k(m)}^0\hQ_h^0(m)\ + \sum_{j=1}^{N_h^k(m)} \a_{N_h^k(m)}^j \Big[ r_h^{\tau_h^j(m)} + \hV_{h+1}^{\tau_h^j(m)}(s_{h+1}^{\tau_h^j(m)}) + \frac{\b}{\sqrt{j}}\Big].
\end{align*}
Using the fact $\sum_{j=0}^t \a_t^j = 1$, and notice that $\a_t^0=0$ for all $t\geq 1$, we have that, as long as $N_h^k(m)\geq 1$,
\begin{align}
\label{eq:optimism-1}
\tQ_h^k(m) - Q^*_h(m) 
=\sum_{j=1}^{N_h^k(m)} \a_{N_h^k(m)}^j \Big[ r_h^{\tau_h^j(m)} + \hV_{h+1}^{\tau_h^j(m)}(s_{h+1}^{\tau_h^j(m)}) - Q^*_h(m) \Big] + \sum_{j=1}^{N_h^k(m)} \a_{N_h^k(m)}^j \cdot\frac{\b}{\sqrt{j}}.
\end{align}
Since the aggregated state representation is $0$-error, there should be
\begin{align}
    Q_h^*(m)
    &= Q_h^*(s_h^{\tau_h^j(m)}, a_h^{\tau_h^j(m)}) \nn\\
    &= r_h^{\tau_h^j(m)} + \tran_h V_{h+1}^*(s_h^{\tau_h^j(m)}, a_h^{\tau_h^j(m)}).
\end{align}
Thus we can rewrite \eqref{eq:optimism-1} as (recall that we assumed deterministic rewards)
\begin{align}
\label{eq:optimism-2}
\tQ_h^k(m) - Q^*_h(m)  
= &\underbrace{\sum_{j=1}^{N_h^k(m)} \a_{N_h^k(m)}^j \cdot\frac{\b}{\sqrt{j}}}_{z_1}\nn\\
    &\ +\underbrace{\sum_{j=1}^{N_h^k(m)} \a_{N_h^k(m)}^j \Big[ \hV_{h+1}^{\tau_h^j(m)}(s_{h+1}^{\tau_h^j(m)}) - V_{h+1}^*(s_{h+1}^{\tau_h^j(m)}) \Big]}_{z_2} \nn\\
    &\ +\underbrace{\sum_{j=1}^{N_h^k(m)} \a_{N_h^k(m)}^j \Big[ V_{h+1}^*(s_{h+1}^{\tau_h^j(m)}) - \tran_h V_{h+1}^*(s_h^{\tau_h^j(m)}, a_h^{\tau_h^j(m)})\Big]}_{z_3}.
\end{align}
Notice that in each summand of $z_3$, there is
\begin{align}   
    \label{eq:zero-mean}
    \tran_h V_{h+1}^*(s_h^{\tau_h^j(m)}, a_h^{\tau_h^j(m)}) = \E\left[V_{h+1}^*(s_{h+1}^{\tau_h^j(m)})\ \Big|\ s_h^{\tau_h^j(m)}, a_h^{\tau_h^j(m)}\right]
\end{align}
Therefore $z_3$ is a martingale difference sequence.  By choosing a specific $\b$, we can make sure that with high probability, $z_1+z_3\geq 0$ uniformly for all $h, k$ and $m$.  Under such event, suppose that additionally we have
\[
    \label{eq:induction-1}
    \hQ_{h+1}^k(m) - Q_{h+1}^*(m) \geq 0,\quad \forall k,m,
\]
then there should be
\[
    \hV_{h+1}^{\tau_h^j(m)}(s_{h+1}^{\tau_h^j(m)}) - V_{h+1}^*(s_{h+1}^{\tau_h^j(m)}) \geq 0,\quad \forall j,m,
\]
which means that $z_2\geq 0$ in \eqref{eq:optimism-2}.  Consequently, $\tQ_h^k(m) - Q^*_h(m)\geq 0$ for all $k$ and $m$.  Recalling that $\hQ_h^k(m)$ is the truncation of $\tQ_h^k(m)$ by $H$, and $Q^*_h(m)\leq H$, we have
\[
    \label{eq:induction-2}
    \hQ_h^k(m) - Q_h^*(m) \geq 0,\quad \forall k,m.
\]
Comparing \eqref{eq:induction-1} and \eqref{eq:induction-2}, we can use backward induction on $h$ to show that the event
\[
    \C{E}_{\rm opt} = \left\{ \hQ_h^k(m) - Q_h^*(m) \geq 0,\ \forall h,k,m\right\}
\]
happens with high probability.

\subsection{Upper Bound for On-Policy Errors}
Now we consider the on-policy error $\hV_h^k(s_h^k) - V_h^*(s_h^k)$. Under the high probability event $\C{E}_{\rm opt}$, the term is apparently lower-bounded by zero.  Here we will provide an upper bound for the sum of the on-policy errors over $k=1,\dots,K$. In fact, since $a_h^k$ is chosen greedily with respect to $\hQ_h^k$, we have that
\[
    \hV_h^k(s_h^k) 
    = \max_{a\in\action} \hQ_h^k(s_h^k, a)=\hQ_h^k(s_h^k, a_h^k).
\]
Also notice that 
\[
    V_h^*(s_h^k) = \max_{a\in\action} Q_h^*(s_h^k, a) \geq Q_h^*(s_h^k, a_h^k).
\]
Using $\htau_h^{j,k}$ and $\hn_h^k$ as shorthands for $\tau_h^j(s_h^k, a_h^k)$ and $N_h^k(s_h^k, a_h^k)$, respectively, we can thus write
\begin{align}
    \label{eq:convergence-1}
    \hV_h^k(s_h^k) - V_h^*(s_h^k) 
    \leq\ &\hQ_h^k(s_h^k, a_h^k) - Q_h^*(s_h^k, a_h^k) \nn\\
    \leq\ &\tQ_h^k(s_h^k, a_h^k) - Q_h^*(s_h^k, a_h^k) \nn\\
    =\ &\sum_{j=1}^{\hn_h^k} \a_{\hn_h^k}^j \left[ r_h^{\htau_h^{j,k}} + \hV_h^{\htau_h^{j,k}}(s_{h+1}^{\htau_h^{j,k}}) + \frac{\b}{\sqrt{j}} \right] - Q_h^*(s_h^k, a_h^k) \nn\\
    =\ &\sum_{j=1}^{\hn_h^k} \a_{\hn_h^k}^j \left[ r_h^{\htau_h^{j,k}} + \hV_h^{\htau_h^{j,k}}(s_{h+1}^{\htau_h^{j,k}}) - Q_h^*(s_h^k, a_h^k)\right] + \sum_{j=1}^{\hn_h^k} \a_{\hn_h^k}^j\cdot\frac{\b}{\sqrt{j}}
\end{align}
By definition, $\htau_h^{j,k}$ is the episode index of the $j$-th visit to aggregate state $\phi_h(s_h^k, a_h^k)$, hence 
\begin{align*}
    Q_h^*(s_h^k, a_h^k) 
    &= Q_h^*(s_h^{\htau_h^{j,k}}, a_h^{\htau_h^{j,k}}) \nn\\
    &= r_h^{\htau_h^{j,k}} + \tran_h V_{h+1}^* (s_h^{\htau_h^{j,k}}, a_h^{\htau_h^{j,k}}),
\end{align*}
where we used the fact that the state-action pairs mapped to one aggregate state share the same ground-truth value, as per the definition of $0$-error aggregation.

Therefore, following \eqref{eq:convergence-1} we have
\begin{align}
    \label{eq:convergence-2}
    \hV_h^k(s_h^k) - V_h^*(s_h^k)
    \leq\ &\hQ_h^k(s_h^k, a_h^k) - Q_h^*(s_h^k, a_h^k) \nn\\
     \leq\ &\underbrace{\sum_{j=1}^{\hn_h^k} \a_{\hn_h^k}^j\cdot\frac{\b}{\sqrt{j}}}_{z'_1} \nn\\
        & + \underbrace{\sum_{j=1}^{\hn_h^k} \a_{\hn_h^k}^j \left[ \hV_{h+1}^{\htau_h^{j,k}}(s_{h+1}^{\htau_h^{j,k}}) - V_{h+1}^*(s_{h+1}^{\htau_h^{j,k}})\right]}_{z'_2} \nn\\
        & + \underbrace{\sum_{j=1}^{\hn_h^k} \a_{\hn_h^k}^j \left[ V_{h+1}^*(s_{h+1}^{\htau_h^{j,k}}) -  \tran_h V_{h+1}^* (s_h^{\htau_h^{j,k}}, a_h^{\htau_h^{j,k}})\right]}_{z'_3}  \nn\\
\end{align}
For the same reason as in \eqref{eq:zero-mean}, $z'_3$ is also a martingale difference sequence. By a proper choice of $\b$, we can ensure that with high probability, uniformly for all $h$ and $k$,
\[
    z'_1 + z'_3 \leq \frac{C}{\sqrt{\hn_h^k}}.
\]
Additionally, the summand $\hV_{h+1}^{\htau_h^{j,k}}(s_{h+1}^{\htau_h^{j,k}}) - V_{h+1}^*(s_{h+1}^{\htau_h^{j,k}})$ in $z'_2$ is the on-policy error at stage $h+1$ of episode $\htau_h^{j,k}$.  As a result, \eqref{eq:convergence-2} can be interpreted as recursively relating the on-policy error at stage $h$ to the on-policy error at stage $h+1$.  By expanding the recursion over $\ell=h,\dots,H$, we can show that there exists $p_1(H,K)$, which is polynomial in $H$ and log-polynomial in $K$, such that
\begin{align}
    \label{eq:on-policy-error}
    \sum_{k=1}^K \hV_h^k(s_h^k) - V_h^*(s_h^k) 
    &\leq \sum_{k=1}^K \hQ_h^k(s_h^k, a_h^k) - Q_h^*(s_h^k, a_h^k) \nn\\
    &\leq p_1(H,K) \cdot \sum_{k=1}^K\sum_{\ell=h}^H \frac{1}{\sqrt{\hn_\ell^k}}.
\end{align}

\subsection{Concluding the Proof}
Under event $\C{E}_{\rm opt}$, we have that
\begin{align}
    \regret(K)
    &= \sum_{k=1}^K V_1^*(s_1) - V_1^{\pi_k}(s_1) \nn\\
    &\leq \sum_{k=1}^K \hV_1^k(s_1) - V_1^{\pi_k}(s_1).
\end{align}
Notice that
\begin{align}
    &\hV_h^k(s_h^k) - V_h^{\pi_k}(s_h^k) \nn\\
    =\ &\hV_h^k(s_h^k) - Q_h^*(s_h^k, a_h^k) 
        + Q_h^*(s_h^k, a_h^k) - V_h^{\pi_k}(s_h^k) \nn\\
    =\ &\underbrace{\hQ_h^k(s_h^k, a_h^k) - Q_h^*(s_h^k, a_h^k)}_{w_1} 
        + \underbrace{Q_h^*(s_h^k, a_h^k) - Q_h^{\pi_k}(s_h^k, a_h^k)}_{w_2}, \nn\\
\end{align}
where we use the fact that $a_h^k$ is selected using $\pi_k$, and that $\pi_k$ is greedy with respect to $\hQ_h^k$.  The sum of $w_1$ terms is already upper bounded in \eqref{eq:on-policy-error}, whereas 
\begin{align}
    \label{eq:decomposition-w2}
    w_2 
    =\ &\tran_h V_{h+1}^*(s_h^k, a_h^k) - \tran_h V_{h+1}^{\pi_k}(s_h^k, a_h^k) \nn\\
    =\ &V_{h+1}^*(s_{h+1}^k) - V_{h+1}^{\pi_k}(s_{h+1}^k) \nn\\
        &+ \left[ \tran_h V_{h+1}^*(s_h^k, a_h^k) - V_{h+1}^*(s_{h+1}^k)\right] \nn\\
        &+ \left[ V_{h+1}^{\pi_k}(s_{h+1}^k) - \tran_h V_{h+1}^{\pi_k}(s_h^k, a_h^k)\right].
\end{align}
If we sum up all $w_2$ terms over $k=1,\dots,K$, the sum of the second and third terms on the right-hand side of \eqref{eq:decomposition-w2} can be upper bounded by Hoeffding inequality (since they are again martingale difference sequences), and the sum of the first term is controlled by \eqref{eq:on-policy-error}.  We can thus arrive at, with high probability,
\[
    \label{eq:regret-upper-bound}
    \regret(K) \leq C\cdot p_1(H,K) \cdot \sum_{k=1}^K\sum_{h=1}^H \frac{1}{\sqrt{\hn_h^k}},
\]
where $C$ is an absolute constant.  Finally, notice that 
\begin{align}
\sum_{k=1}^K\sum_{h=1}^H \frac{1}{\sqrt{\dn_h^k}}
    &= \sum_{h=1}^H\sum_{m=1}^M\sum_{j=1}^{N_h^K(m)} \frac{1}{\sqrt{j}} \nn\\
    &\leq \sum_{h=1}^H\sum_{m=1}^M 2\sqrt{N_h^K(m)} \nn\\
    &\leq 2\sqrt{HM\cdot \sum_{h=1}^H\sum_{m=1}^M N_h^K(m)}\label{eq:cauchy}\\
    &= 2\sqrt{H^2MK}, \label{eq:sum-of-inverse-square-roots}
\end{align}
where \eqref{eq:cauchy} comes from Cauchy-Schwartz inequality and \eqref{eq:sum-of-inverse-square-roots} is because there are $HK$ stages in total.  Combining \eqref{eq:regret-upper-bound} and \eqref{eq:sum-of-inverse-square-roots}, we arrive at the desired result.

\bibliography{bibliography_icml}
\bibliographystyle{icml2020}

\newpage
\appendix

\section{Proof of Theorem \ref{thm:main}}
We now provide the complete proof of Theorem \ref{thm:main}.  We will no longer assume that $\{\phi_h\}_{h\in[H]}$ is a sufficient aggregation.  In the following, let $\{\phi_h\}_{h\in[H]}$ be an $\e$-error aggregation, with $\e\geq 0$.  Throughout we will use the same set of notations as in the main body of the paper: 
\begin{itemize}
	\item $\hQ_h^k(m)$: the value function estimate $\hQ_h$ of aggregate state $m$, at the end of episode $k$, with $\hQ_h^0(m) = H$.
	\item $\tQ_h^k(m)$: the uncapped value function estimate $\tQ_h$ of aggregate state $m$, at the end of episode $k$.  We have that
	\[
		\hQ_h^k(m) = \min\left\{\tQ_h^k(m), H \right\}.
	\]
	\item $N_h^k(m)$: the number of visits to aggregate state $m$ at stage $h$ in the first $k$ trajectories (indexed from 0 to $k-1$).
	\item $\tau_h^j(m)$: the episode index of the $j$-th visit to aggregate state $m$, at stage $h$. There should be
    \[
        \phi_h(s_h^{\tau_h^j(m)}, a_h^{\tau_h^j(m)}) = m,\quad \forall j=1,2\dots
    \]
	\item $\pi_k$: the greedy policy with respect to $\{\hQ_h^k(m)\}$, i.e. the policy that the agent follows to produce the trajectory $s_1^k, a_1^k, s_2^k, \dots, s_H^k, a_H^k$.  The final policy output by the algorithm is $\pi_K$.
	\item $\hV_h^k(s)$: the state value function estimate at stage $h$, induced by $\hQ_h^k(m)$ through
	\[
		\hV_h^k(s) = \max_{a\in\action} \hQ_h^k(\phi_h(s, a)).\nn
	\] 
\end{itemize}
We will also use the simplified notations $\hQ_h^k(s, a)$, $\tQ_h^k(s, a)$, $N_h^k(s, a)$ and $\tau_h^j(s,a)$to represent $\hQ_h^k(\phi_h(s, a))$, $\tQ_h^k(\phi_h(s, a))$, $N_h^k(\phi_h(s, a))$ and $\tau_h^j(\phi_h(s,a))$, respectively. Note that we will stop using the $Q^*(m)$ notation since state-action pairs mapped to one aggregate state can have different ground-truth values.  To take into account the dynamics of the final stage $H$, when the subscript is $H+1$, we define
\[
	\tQ_{H+1}^k(s,a) = \hQ_{H+1}^k(s,a) = \hV_{H+1}^k(s) = Q_{H+1}^*(s,a) = V_{H+1}^*(s) = 0, \quad\forall s, a, k.
\]
The $\{\b_i\}$ sequence, which controls the amount of optimism boost added to the value estimates, is
\[
        \b_i = 2H^{\frac{3}{2}}\sqrt{\log\frac{HK}{\d}} + \e\cdot \sqrt{i},\quad i=1,2\dots,
\]
Also recall that the sequence of stepsizes are
\[
	\a_t = \frac{H+1}{H+t},\quad t=1,2,\dots,
\]
and we adopt the notations 
\[
	\a_t^0 = \prod_{j=1}^t (1 - \a_j)
\]
and 
\[
	\a_t^i = \a_i\prod_{j=i + 1}^t (1 - \a_j).
\]
The following lemma from \cite{jin2018q} demonstrates the desirable properties with respect to this specific sequence of stepsizes.
\begin{lem}
    \label{lem:stepsize}
    {\bf (Lemma 4.1 in \cite{jin2018q})} We have that
    \begin{enumerate}[label=(\alph*)]
        \item For every $t \geq 1$, $\frac{1}{\sqrt{t}} \leq \sum_{i=1}^t \frac{\a_t^i}{\sqrt{i}} \leq \frac{2}{\sqrt{t}}$;
        \item For every $t \geq 1$, $\max_{i\in[t]} \a_t^i \leq \frac{2H}{t}$ and $\sum_{i=1}^t (\a_t^i)^2 \leq \frac{2H}{t}$;
        \item For every $i\geq 1$, $\sum_{t=i}^\infty \a_t^i = 1 + \frac{1}{H}$.
    \end{enumerate}
\end{lem}
The rest of this supplementary material is organized as follows.  In Subsection \ref{sec:optimism} we show that the value function estimates are optimistic with high probability; in Subsection \ref{sec:upper-bound} we establish a polynomial upper bound for the on-policy errors; the proof is concluded in \ref{sec:concluding}.

\subsection{Optimism}
\label{sec:optimism}
Recall that the uncapped value functions estimates can be written as
\begin{align*}
\tQ_h^k(m)
=\a_{N_h^k(m)}^0\hQ_h^0(m)\ + 
 \sum_{j=1}^{N_h^k(m)} \a_{N_h^k(m)}^j \Big[ r_h^{\tau_h^j(m)} + \hV_{h+1}^{\tau_h^j(m)}(s_{h+1}^{\tau_h^j(m)}) + \frac{\b_j}{\sqrt{j}}\Big].
\end{align*}
Therefore we have that
\[
    \tQ_h^k(s, a) - Q^*_h(s, a)
    &=& \a_{N_h^k(s,a)}^0\cdot \big(H - Q^*_h(s, a)\big) + \nn\\
        &&\sum_{j=1}^{N_h^k(s,a)}\a_{N_h^k(s,a)}^j \left[ r_h^{\tau_h^j(s,a)} + \h{V}_{h+1}^{\tau_h^j(s,a)}(s_{h+1}^{\tau_h^j(s,a)}) + \frac{\b_j}{\sqrt{j}} - Q_h^*(s,a)\right] \label{eq:1}\\
    &=& \a_{N_h^k(s,a)}^0\cdot \big(H - Q^*_h(s, a)\big) + \nn\\
        && \sum_{j=1}^{N_h^k(s,a)}\a_{N_h^k(s,a)}^j \left[ r_h^{\tau_h^j(s,a)} + \h{V}_{h+1}^{\tau_h^j(s,a)}(s_{h+1}^{\tau_h^j(s,a)}) + \frac{\b_j}{\sqrt{j}} - Q_h^*(s_h^{\tau_h^j(s,a)},a_h^{\tau_h^j(s,a)})\right] + \nn\\
        && \sum_{j=1}^{N_h^k(s,a)}\a_{N_h^k(s,a)}^j \left[Q_h^*(s_h^{\tau_h^j(s,a)},a_h^{\tau_h^j(s,a)}) - Q_h^*(s,a)\right]\nn\\
    &\geq& \a_{N_h^k(s,a)}^0\cdot \big(H - Q^*_h(s, a)\big) -\e + \nn\\
        && \sum_{j=1}^{N_h^k(s,a)}\a_{N_h^k(s,a)}^j \left[ r_h^{\tau_h^j(s,a)} + \h{V}_{h+1}^{\tau_h^j(s,a)}(s_{h+1}^{\tau_h^j(s,a)}) + \frac{\b_j}{\sqrt{j}} - Q_h^*(s_h^{\tau_h^j(s,a)},a_h^{\tau_h^j(s,a)})\right] \label{eq:2}\\
    &\geq& \a_{N_h^k(s,a)}^0\cdot \big(H - Q^*_h(s, a)\big) - \e + \nn\\
        &&\sum_{j=1}^{N_h^k(s,a)}\a_{N_h^k(s,a)}^j \left[  \h{V}_{h+1}^{\tau_h^j(s,a)}(s_{h+1}^{\tau_h^j(s,a)}) - V_{h+1}^*(s_{h+1}^{\tau_h^j(s,a)})\right] +\nn\\
        &&\sum_{j=1}^{N_h^k(s,a)}\a_{N_h^k(s,a)}^j \left[ \frac{\b_j}{\sqrt{j}} +V_{h+1}^*(s_{h+1}^{\tau_h^j(s,a)}) -  \tran_h V_{h+1}^*(s_h^{\tau_h^j(s,a)}, a_h^{\tau_h^j(s,a)})\right], \label{eq:3}
\]
where we used the fact that $\sum_{j=0}^t\a_t^j = 1$ in \eqref{eq:1}.  Inequality \eqref{eq:2} results from that
\[
	\left| Q_h^*(s,a) - Q_h^*(s_h^{\tau_h^j(s,a)},a_h^{\tau_h^j(s,a)}) \right|
	\leq \e,
\]
which is the consequence of 
\[
	\phi_h(s,a) = \phi_h(s_h^{\tau_h^j(s,a)},a_h^{\tau_h^j(s,a)}),
\]
and that $\phi_h$ is an $\e$-error aggregation.  The final step \eqref{eq:3} is from the property of the ground-truth value function:
\[
	Q_h^*(s', a') = R_h(s', a') + \tran_h V_{h+1}^*(s', a'),\quad \forall (s', a')\in\state\times\action.
\]
By Azuma-Hoeffding inequality, with probability at least $1-\d$, for all $h\in[H]$ and $k\in[K]$,
\[
    \label{eq:azuma-hoeffding-1}
    \left|\sum_{j=1}^{N_h^k(s,a)}\a_{N_h^k(s,a)}^j \left[ V_{h+1}^*(s_{h+1}^{\tau_h^j(s,a)}) -  \tran_h V_{h+1}^*(s_h^{\tau_h^j(s,a)}, a_h^{\tau_h^j(s,a)})\right]\right|
    \geq 
    -\frac{2H^{\frac{3}{2}}}{\sqrt{N_h^k(s,a)}}\cdot \sqrt{\log\frac{HK}{\d}}.
\]
We denote the above event as $\C{E}_{\rm opt}$. 
Notice that
\[
	\a_t^0 = 
	\begin{cases}
		1 & \text{if }t = 0\\
		0 & \text{otherwise}
	\end{cases}.
\]
Therefore, under event $\C{E}_{\rm opt}$, whenever $N_h^k(s,a) \geq 1$, we have
\[
	\tQ_h^k(s, a) - Q^*_h(s, a)
	&\geq& \sum_{j=1}^{N_h^k(s,a)}\a_{N_h^k(s,a)}^j \left[  \h{V}_{h+1}^{\tau_h^j(s,a)}(s_{h+1}^{\tau_h^j(s,a)}) - V_{h+1}^*(s_{h+1}^{\tau_h^j(s,a)})\right]\nn\\
		&& - \frac{2H^{\frac{3}{2}}}{\sqrt{N_h^k(s,a)}}\cdot \sqrt{\log\frac{HK}{\d}} + \left\{ \sum_{j=1}^{N_h^k(s,a)}\a_{N_h^k(s,a)}^j \cdot \frac{\b_j}{\sqrt{j}} - \e \right\} \nn\\
	&=& \sum_{j=1}^{N_h^k(s,a)}\a_{N_h^k(s,a)}^j \left[  \h{V}_{h+1}^{\tau_h^j(s,a)}(s_{h+1}^{\tau_h^j(s,a)}) - V_{h+1}^*(s_{h+1}^{\tau_h^j(s,a)})\right]\nn\\
		&&- \frac{2H^{\frac{3}{2}}}{\sqrt{N_h^k(s,a)}}\cdot \sqrt{\log\frac{HK}{\d}} + \left\{ \sum_{j=1}^{N_h^k(s,a)}\a_{N_h^k(s,a)}^j \cdot \left(\frac{\b_j}{\sqrt{j}} - \e\right) \right\} \nn\\
	&=& \sum_{j=1}^{N_h^k(s,a)}\a_{N_h^k(s,a)}^j \left[  \h{V}_{h+1}^{\tau_h^j(s,a)}(s_{h+1}^{\tau_h^j(s,a)}) - V_{h+1}^*(s_{h+1}^{\tau_h^j(s,a)})\right]\nn\\
		&&- \frac{2H^{\frac{3}{2}}}{\sqrt{N_h^k(s,a)}}\cdot \sqrt{\log\frac{HK}{\d}} + 2H^{\frac{3}{2}}\sqrt{\log\frac{HK}{\d}}\cdot \sum_{j=1}^{N_h^k(s,a)}\frac{\a_{N_h^k(s,a)}^j}{\sqrt{j}}\nn\\
	&\geq& \sum_{j=1}^{N_h^k(s,a)}\a_{N_h^k(s,a)}^j \left[  \h{V}_{h+1}^{\tau_h^j(s,a)}(s_{h+1}^{\tau_h^j(s,a)}) - V_{h+1}^*(s_{h+1}^{\tau_h^j(s,a)})\right], \label{eq:4}
\]
where the final step comes from (a) in Lemma \ref{lem:stepsize}.  When $N_h^k(s,a) = 0$, there is
\[
	\tQ_h^k(s, a) = \hQ_h^k(s, a) = H.
\]
Plugging $h=H$ into \eqref{eq:4}, we have $\tQ_H^k(s, a) \geq Q^*_H(s, a)$ for all $s$ and $a$. 
Considering that $\hQ_H^k(s, a) = \tQ^*_H(s, a)$ whenever $\tQ^*_H(s, a)\leq H$ and $Q^*_H(s, a) \leq H$, we have that
\[
    \hQ_H^k(s, a) \geq Q^*_H(s, a),\quad \forall (s,a)\in\state\times\action, k\in[K].
\]
Therefore,
\[
    \hV_H^k(s) \geq V^*_H(s),\quad \forall s\in\state, k\in[K].
\]
Suppose that at stage $h$,
\[
    \hV_h^k(s) \geq V^*_h(s),\quad \forall s\in\state, k\in[K].
\]
Then from \eqref{eq:4},
\[
    \tQ_{h-1}^k(s, a) - Q^*_{h-1}(s, a) \geq \sum_{j=1}^{N_h^k(s,a)}\a_{N_h^k(s,a)}^j \left[  \h{V}_{h+1}^{\tau_h^j(s,a)}(s_{h+1}^{\tau_h^j(s,a)}) - V_{h+1}^*(s_{h+1}^{\tau_h^j(s,a)})\right] \geq 0.
\]
This leads to
\[
    \hQ_{h-1}^k(s, a) \geq Q^*_{h-1}(s, a),\quad \forall (s,a)\in\state\times\action, k\in[K],
\]
and further, by maximizing over $a\in\action$ on both sides,
\[
    \label{eq:induction-end}
    \hV_{h-1}^k(s) \geq V^*_{h-1}(s),\quad \forall s\in\state, k\in[K].
\]
Therefore, by the induction principle, we have that under $\C{E}_{\rm opt}$,
\[
    \label{eq:optimism-theorem-1}
    \hV_h^k(s) \geq V^*_h(s),\quad \forall s\in\state, h\in[H], k\in[K].
\]
Thus we have shown that the value function estimates are optimistic with high probability.

\subsection{Upper Bound for On-policy Errors}
\label{sec:upper-bound}
Under event $\C{E}_{\rm opt}$, the on-policy error $\hV_h^k(s_h^k) - V_h^*(s_h^k)$ is lower bounded by zero.  We also have that
\begin{align}
	\hV_h^k(s_h^k) - V^*_h(s_h^k)
    &\leq  \hQ_h^k(s_h^k, a_h^k) - Q^*_h(s_h^k, a_h^k) \label{eq:1} \\
    &\leq  \tQ_h^k(s_h^k, a_h^k) - Q^*_h(s_h^k, a_h^k) \nn \\
    &\leq \a_{N_h^k(s_h^k, a_h^k)}^0\cdot \big(H - Q^*_h(s_h^k, a_h^k)\big)  \nn\\
        &\ +\sum_{j=1}^{N_h^k(s_h^k, a_h^k)}\a_{N_h^k(s_h^k, a_h^k)}^j \left[ r_h^{\tau_h^j(s_h^k, a_h^k)} + \h{V}_{h+1}^{\tau_h^j(s_h^k, a_h^k)}(s_{h+1}^{\tau_h^j(s_h^k, a_h^k)}) + \frac{\b}{\sqrt{j}} - Q^*_h(s_h^k, a_h^k) \right] \nn\\
    &= \a_{N_h^k(s_h^k, a_h^k)}^0\cdot \big(H - Q^*_h(s_h^k, a_h^k)\big)  \nn\\
        &\ +\sum_{j=1}^{N_h^k(s_h^k, a_h^k)}\a_{N_h^k(s_h^k, a_h^k)}^j \left[ r_h^{\tau_h^j(s_h^k, a_h^k)} + \h{V}_{h+1}^{\tau_h^j(s_h^k, a_h^k)}(s_{h+1}^{\tau_h^j(s_h^k, a_h^k)}) + \frac{\b}{\sqrt{j}} - Q^*_h(s_h^{\tau_h^j(s_h^k, a_h^k)}, a_h^{\tau_h^j(s_h^k, a_h^k)}) \right]  \nn\\
        &\ +\sum_{j=1}^{N_h^k(s_h^k, a_h^k)}\a_{N_h^k(s_h^k, a_h^k)}^j \left[ Q^*_h(s_h^{\tau_h^j(s_h^k, a_h^k)}, a_h^{\tau_h^j(s_h^k, a_h^k)}) - Q^*_h(s_h^k, a_h^k) \right] \nn\\
    &\leq \a_{N_h^k(s_h^k, a_h^k)}^0\cdot \big(H - Q^*_h(s_h^k, a_h^k)\big) + \e  \nn\\
        &\ +\sum_{j=1}^{N_h^k(s_h^k, a_h^k)}\a_{N_h^k(s_h^k, a_h^k)}^j \left[ r_h^{\tau_h^j(s_h^k, a_h^k)} + \h{V}_{h+1}^{\tau_h^j(s_h^k, a_h^k)}(s_{h+1}^{\tau_h^j(s_h^k, a_h^k)}) + \frac{\b}{\sqrt{j}} - Q^*_h(s_h^{\tau_h^j(s_h^k, a_h^k)}, a_h^{\tau_h^j(s_h^k, a_h^k)}) \right] \nn\\
    &= \a_{N_h^k(s_h^k, a_h^k)}^0\cdot \big(H - Q^*_h(s_h^k, a_h^k)\big)   \nn\\
        &\ +\underbrace{\sum_{j=1}^{N_h^k(s_h^k, a_h^k)}\a_{N_h^k(s_h^k, a_h^k)}^j \Bigg[  \h{V}_{h+1}^{\tau_h^j(s_h^k, a_h^k)}(s_{h+1}^{\tau_h^j(s_h^k, a_h^k)}) - V_{h+1}^*(s_{h+1}^{\tau_h^j(s_h^k, a_h^k)})\Bigg]}_{q_1}  \nn\\
        &\ +\underbrace{\sum_{j=1}^{N_h^k(s_h^k, a_h^k)}\a_{N_h^k(s_h^k, a_h^k)}^j \Bigg[V_{h+1}^*(s_{h+1}^{\tau_h^j(s_h^k, a_h^k)}) - \tran_h V_{h+1}^*(s_h^{\tau_h^j(s_h^k, a_h^k)}, a_h^{\tau_h^j(s_h^k, a_h^k)})\Bigg]}_{q_2} \nn\\
        &\ +\underbrace{\e + \sum_{j=1}^{N_h^k(s_h^k, a_h^k)}\a_{N_h^k(s_h^k, a_h^k)}^j \frac{\b}{\sqrt{j}} }_{q_3},\label{eq:5}
\end{align}
where we used the fact that 
\[
	\phi_h(s_h^{\tau_h^j(s_h^k, a_h^k)}, a_h^{\tau_h^j(s_h^k, a_h^k)}) = \phi_h(s_h^k, a_h^k),
\]
which leads to 
\[
	\left| Q^*_h(s_h^{\tau_h^j(s_h^k, a_h^k)}, a_h^{\tau_h^j(s_h^k, a_h^k)}) - Q^*_h(s_h^k, a_h^k) \right| \leq \e.
\]
The final equality \eqref{eq:5} is because
\[
	Q^*_h(s_h^{\tau_h^j(s_h^k, a_h^k)}, a_h^{\tau_h^j(s_h^k, a_h^k)})
	=
	r_h^{\tau_h^j(s_h^k, a_h^k)} + \tran_h V_{h+1}^*(s_h^{\tau_h^j(s_h^k, a_h^k)}, a_h^{\tau_h^j(s_h^k, a_h^k)})
\]
by definition.  Notice that under event $\C{E}_{\rm opt}$, there is 
\[
	\label{eq:q-2}
	q_2 \leq \frac{2H^{\frac{3}{2}}}{\sqrt{N_h^k(s_h^k,a_h^k)}}\cdot \sqrt{\log\frac{HK}{\d}}.
\]
In addition, whenever $N_h^k(s_h^k,a_h^k) \geq 1$, there is
\[
	\label{eq:q-3}
	q_3
	&=& \sum_{j=1}^{N_h^k(s_h^k, a_h^k)}\a_{N_h^k(s_h^k, a_h^k)}^j \left( \frac{\b_j}{\sqrt{j}} + \e\right) \nn\\
	&=& 2\e + 2H^{\frac{3}{2}}\sqrt{\log\frac{HK}{\d}}\cdot \sum_{j=1}^{N_h^k(s_h^k, a_h^k)} \frac{\a_{N_h^k(s_h^k, a_h^k)}^j}{\sqrt{j}}\nn\\
	&\leq& 2\e + \frac{4H^{\frac{3}{2}}}{\sqrt{N_h^k(s_h^k, a_h^k)}} \cdot \sqrt{\log\frac{HK}{\d}}.
\]
Let 
\[
	\chi_h^k = \hV_h^k(s_h^k) - V^*_h(s_h^k)
\]
be the on-policy error. 
Combining \eqref{eq:5}, \eqref{eq:q-2} and \eqref{eq:q-3}, we arrive at, whenever $N_h^k(s_h^k,a_h^k) \geq 1$,
\[
	\label{eq:6}
	\chi_h^k
	&\leq& \hQ_h^k(s_h^k, a_h^k) - Q^*_h(s_h^k, a_h^k) \nn\\
	&\leq& \sum_{j=1}^{N_h^k(s_h^k, a_h^k)}\a_{N_h^k(s_h^k, a_h^k)}^j \cdot \chi_{h+1}^{\tau_h^j(s_h^k, a_h^k)}  + \frac{6H^{\frac{3}{2}}}{\sqrt{N_h^k(s_h^k, a_h^k)}} \cdot \sqrt{\log\frac{HK}{\d}} + 2\e.
\]
Notice that \eqref{eq:6} is recursive, in that it relates the on-policy error at stage $h$ to the on-policy error at stage $h+1$. Summing both sides over $k=1,\dots,K$, we have
\[
	\label{eq:7}
	\sum_{k=1}^K \chi_h^k
	&\leq& \sum_{k=1}^K \hQ_h^k(s_h^k, a_h^k) - Q^*_h(s_h^k, a_h^k)\nn\\
	&\leq& \frac{6H^{\frac{3}{2}}K}{\sqrt{N_h^k(s_h^k, a_h^k)}} \cdot \sqrt{\log\frac{HK}{\d}} + 2\e K + \sum_{k=1}^K\sum_{j=1}^{N_h^k(s_h^k, a_h^k)}\a_{N_h^k(s_h^k, a_h^k)}^j \cdot \chi_{h+1}^{\tau_h^j(s_h^k, a_h^k)} .
\]
Notice that
\[
	\label{eq:8}
    \sum_{k=1}^K\sum_{j=1}^{N_h^k(s_h^k, a_h^k)} \a_{N_h^k(s_h^k, a_h^k)}^j \cdot \chi_{h+1}^{\tau_h^j(s_h^k, a_h^k)}
    \leq \sum_{k=1}^K \chi_{h+1}^k \cdot \sum_{t=N_h^k(s_h^k, a_h^k)+1}^\infty \a_t^{N_h^k(s_h^k, a_h^k)}
    \leq \left(1 + \frac{1}{H}\right)\sum_{k=1}^K \chi_{h+1}^k,
\]
where the final inequality results from (c) in Lemma \ref{lem:stepsize}.  Hence \eqref{eq:7} becomes 
\[
	\label{eq:9}
	\sum_{k=1}^K \chi_h^k
	&\leq& \sum_{k=1}^K \hQ_h^k(s_h^k, a_h^k) - Q^*_h(s_h^k, a_h^k)\nn\\
	&\leq& \frac{6H^{\frac{3}{2}}K}{\sqrt{N_h^k(s_h^k, a_h^k)}} \cdot \sqrt{\log\frac{HK}{\d}} + 2\e K + \left(1 + \frac{1}{H}\right)\sum_{k=1}^K \chi_{h+1}^k.
\]
Expanding the recursion \eqref{eq:8} over stages $h, h+1,\dots, H$, we have that, Under event $\C{E}_{\rm opt}$,
\[
	\label{eq:10}
	\sum_{k=1}^K \chi_h^k
	&\leq& \sum_{k=1}^K \hQ_h^k(s_h^k, a_h^k) - Q^*_h(s_h^k, a_h^k)\nn\\
	&\leq& \left( 1 + \frac{1}{H} \right)^H \cdot \left\{2\e HK + \sum_{\ell=h}^H\frac{6H^{\frac{5}{2}}K}{\sqrt{N_\ell^k(s_\ell^k, a_\ell^k)}} \cdot \sqrt{\log\frac{HK}{\d}}  \right\} \nn\\
	&\leq& 6\e HK + 18 H^{\frac{5}{2}}K \sqrt{\log\frac{HK}{\d}} \cdot \sum_{\ell=h}^H \frac{1}{\sqrt{N_\ell^k(s_\ell^k, a_\ell^k)}}.
\]

\subsection{Concluding the Proof}
\label{sec:concluding}
Under event $\C{E}_{\rm opt}$, we have 
\begin{eqnarray}
	\label{eq:regret-decomposition}
    \regret(K)
    &=& \sum_{k=1}^K V_1^*(s_1) - V_1^{\pi_k}(s_1) \nn\\
    &\leq& \sum_{k=1}^K \hV_1^k(s_1) - V_1^{\pi_k}(s_1).
\end{eqnarray}
On the other hand, under event $\C{E}_{\rm opt}$,
\[
	\label{eq:11}
    \hV_h^k(s_h^k) - V_h^{\pi_k}(s_h^k)
    &=& \hV_h^k(s_h^k) - Q_h^*(s_h^k, a_h^k) 
        + Q_h^*(s_h^k, a_h^k) - V_h^{\pi_k}(s_h^k) \nn\\
    &=& \hQ_h^k(s_h^k, a_h^k) - Q_h^*(s_h^k, a_h^k)
        + Q_h^*(s_h^k, a_h^k) - Q_h^{\pi_k}(s_h^k, a_h^k) \nn\\
    &=& \hQ_h^k(s_h^k, a_h^k) - Q_h^*(s_h^k, a_h^k) + \tran_h V_{h+1}^*(s_h^k, a_h^k) - \tran_h V_{h+1}^{\pi_k}(s_h^k, a_h^k) \nn\\
    &=& \left[ \hQ_h^k(s_h^k, a_h^k) - Q_h^*(s_h^k, a_h^k) \right] + \left[ V_{h+1}^*(s_{h+1}^k) - V_{h+1}^{\pi_k}(s_{h+1}^k) \right] \nn\\
    	&&+ \left[ \tran_h V_{h+1}^*(s_h^k, a_h^k) - V_{h+1}^*(s_{h+1}^k) \right] + \left[ V_{h+1}^{\pi_k}(s_{h+1}^k) - \tran_h V_{h+1}^{\pi_k}(s_h^k, a_h^k)\right] \nn\\
    &=& \left[ \hQ_h^k(s_h^k, a_h^k) - Q_h^*(s_h^k, a_h^k) \right] + \left[ \hV_{h+1}^k(s_{h+1}^k) - V_{h+1}^{\pi_k}(s_{h+1}^k) \right] - \chi_{h+1}^k \nn\\
    	&&+ \left[ \tran_h V_{h+1}^*(s_h^k, a_h^k) - V_{h+1}^*(s_{h+1}^k) \right] + \left[ V_{h+1}^{\pi_k}(s_{h+1}^k) - \tran_h V_{h+1}^{\pi_k}(s_h^k, a_h^k)\right].
\]
Summing over $k=1,\dots, K$, and take into account \eqref{eq:8} and \eqref{eq:10}, we have
\[
	\label{eq:12}
	\sum_{k=1}^K \hV_h^k(s_h^k) - V_h^{\pi_k}(s_h^k)
	&\leq& \sum_{k=1}^K \left[ \hV_{h+1}^k(s_{h+1}^k) - V_{h+1}^{\pi_k}(s_{h+1}^k) \right] - \sum_{k=1}^K \chi_{h+1}^k \nn\\
		&& + \left(1 + \frac{1}{H}\right)\sum_{k=1}^K \chi_{h+1}^k  + \frac{6H^{\frac{3}{2}}K}{\sqrt{N_h^k(s_h^k, a_h^k)}} \cdot \sqrt{\log\frac{HK}{\d}} + 2\e K \nn\\
		&& + \sum_{k=1}^K \left[ \tran_h V_{h+1}^*(s_h^k, a_h^k) - V_{h+1}^*(s_{h+1}^k) \right] \nn\\
		&& + \sum_{k=1}^K \left[ V_{h+1}^{\pi_k}(s_{h+1}^k) - \tran_h V_{h+1}^{\pi_k}(s_h^k, a_h^k)\right].
\]
Notice that
\[
	\chi_{h+1}^k = \hV_{h+1}^k(s_{h+1}^k) - V_{h+1}^k(s_{h+1}^k) \leq \hV_{h+1}^k(s_{h+1}^k) - V_{h+1}^{\pi_k}(s_{h+1}^k).
\]
Hence \eqref{eq:12} can be rewritten as
\[
	\label{eq:13}
	\sum_{k=1}^K \hV_h^k(s_h^k) - V_h^{\pi_k}(s_h^k)
	&\leq& \left(1 + \frac{1}{H}\right)\sum_{k=1}^K \left[ \hV_{h+1}^k(s_{h+1}^k) - V_{h+1}^{\pi_k}(s_{h+1}^k) \right] \nn\\
		&& +\ 2\e K + 6H^{\frac{3}{2}}\cdot \sqrt{\log\frac{HK}{\d}}\sum_{k=1}^K\frac{1}{\sqrt{N_h^k(s_h^k, a_h^k)}}  \nn\\
		&& + \sum_{k=1}^K \left[ \tran_h V_{h+1}^*(s_h^k, a_h^k) - V_{h+1}^*(s_{h+1}^k) \right] \nn\\
		&& + \sum_{k=1}^K \left[ V_{h+1}^{\pi_k}(s_{h+1}^k) - \tran_h V_{h+1}^{\pi_k}(s_h^k, a_h^k)\right].
\]
The above inequality is also recursive.  We thus have (recall that $s_1^k=s_1$)
\[
	\label{eq:14}
	\sum_{k=1}^K \hV_1^k(s_1) - V_1^{\pi_k}(s_1)
	&\leq& 6\e HK + 18H^{\frac{3}{2}}\cdot \sqrt{\log\frac{HK}{\d}}\sum_{h=1}^H\sum_{k=1}^K\frac{1}{\sqrt{N_h^k(s_h^k, a_h^k)}}  \nn\\
		&& + 3\sum_{h=1}^H\sum_{k=1}^K \left[ \tran_h V_{h+1}^*(s_h^k, a_h^k) - V_{h+1}^*(s_{h+1}^k) \right] \nn\\
		&& + 3\sum_{h=1}^H\sum_{k=1}^K \left[ V_{h+1}^{\pi_k}(s_{h+1}^k) - \tran_h V_{h+1}^{\pi_k}(s_h^k, a_h^k)\right].
\]
Meanwhile, there is also
\[
	\sum_{k=1}^K \left[ \tran_h V_{h+1}^*(s_h^k, a_h^k) - V_{h+1}^*(s_{h+1}^k) \right]
	&=& \sum_{m=1}^M \sum_{k=1}^K  \ind\left\{ \phi_h(s_h^k, a_h^k) = m\right\} \cdot \left[ \tran_h V_{h+1}^*(s_h^k, a_h^k) - V_{h+1}^*(s_{h+1}^k) \right].
\]
By Azuma-Hoeffding inequality, with probability at least $1-\d$, 
\[
	\sum_{k=1}^K  \ind\left\{ \phi_h(s_h^k, a_h^k) = m\right\} \cdot \left[ \tran_h V_{h+1}^*(s_h^k, a_h^k) - V_{h+1}^*(s_{h+1}^k) \right]
	\leq \frac{2H^{\frac{3}{2}}}{\sqrt{N_h^K(m)}} \cdot \sqrt{\log\frac{HK}{\d}}.
\]
Hence with probability at least $1-2\d$,
\[
	\label{eq:15}
	\sum_{k=1}^K \left[ \tran_h V_{h+1}^*(s_h^k, a_h^k) - V_{h+1}^*(s_{h+1}^k) \right] + \sum_{k=1}^K \left[ V_{h+1}^{\pi_k}(s_{h+1}^k) - \tran_h V_{h+1}^{\pi_k}(s_h^k, a_h^k)\right] \leq 4\sum_{m=1}^M \frac{2H^{\frac{3}{2}}}{\sqrt{N_h^K(m)}} \cdot \sqrt{\log\frac{HK}{\d}}.
\]
Finally, we have 
\[
	\label{eq:16}
	\sum_{k=1}^K\sum_{h=1}^H \frac{1}{\sqrt{N_h^k(s_h^k, a_h^k)}}
    &=& \sum_{h=1}^H\sum_{m=1}^M\sum_{j=1}^{N_h^K(m)} \frac{1}{\sqrt{j}} \nn\\
    &\leq& \sum_{h=1}^H\sum_{m=1}^M 2\sqrt{N_h^K(m)} \nn\\
    &\leq& 2\sqrt{HM\cdot \sum_{h=1}^H\sum_{m=1}^M N_h^K(m)} = 2\sqrt{H^2MK}.
\]
Combining \eqref{eq:regret-decomposition}, \eqref{eq:14}, \eqref{eq:15}, and \eqref{eq:16}, and scaling $\d$ to $\d/3$, we arrive at the desired result.

\end{document}